\documentclass{bmvc2k}

\usepackage{bm}

\title{Selective Scene Text Removal}

\addauthor{Hayato Mitani}{hayato.mitani@human.ait.kyushu-u.ac.jp}{1}
\addauthor{Akisato Kimura}{akisato@ieee.org}{2}
\addauthor{Seiichi Uchida}{uchida@ait.kyushu-u.ac.jp}{1}

\addinstitution{
 Kyushu University,\\
 Fukuoka, Japan
}
\addinstitution{
 NTT Corporation,\\
 Kanagawa, Japan
}

\runninghead{Mitani, Kimura, and Uchida}{Selective Scene Text Removal}

\begin{document}

\maketitle

\begin{abstract}
Scene text removal (STR) is the image transformation task to remove text regions in scene images. The conventional STR methods remove all scene text. This means that the existing methods cannot select text to be removed. In this paper, we propose a novel task setting named selective scene text removal (SSTR) that removes only target words specified by the user. Although SSTR is a more complex task than STR, the proposed multi-module structure enables efficient training for SSTR. Experimental results show that the proposed method can remove target words as expected.
\end{abstract}

\section{Introduction\label{sec:intro}}
\begin{figure}[t]
 \begin{center}
 \includegraphics[keepaspectratio, width=0.75\linewidth]{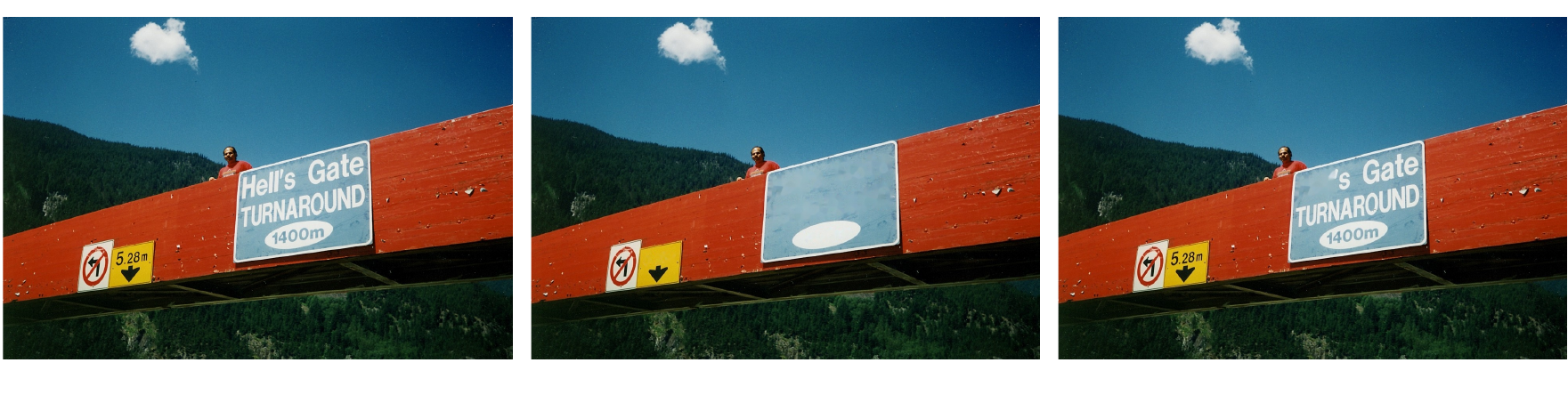}\\[-5mm]
\caption{The difference between the conventional scene text removal (STR) and our selective scene text removal (SSTR). From left to right, the original image, STR to remove all texts, SSTR to remove only a target word {\tt Hell}. Note that these STR and SSTR images are artificial examples. 
{\small (The example image is taken from Flickr Creative Commons, public domain.)}}  \label{fig:task}
\smallskip
 \includegraphics[keepaspectratio, width=\linewidth]{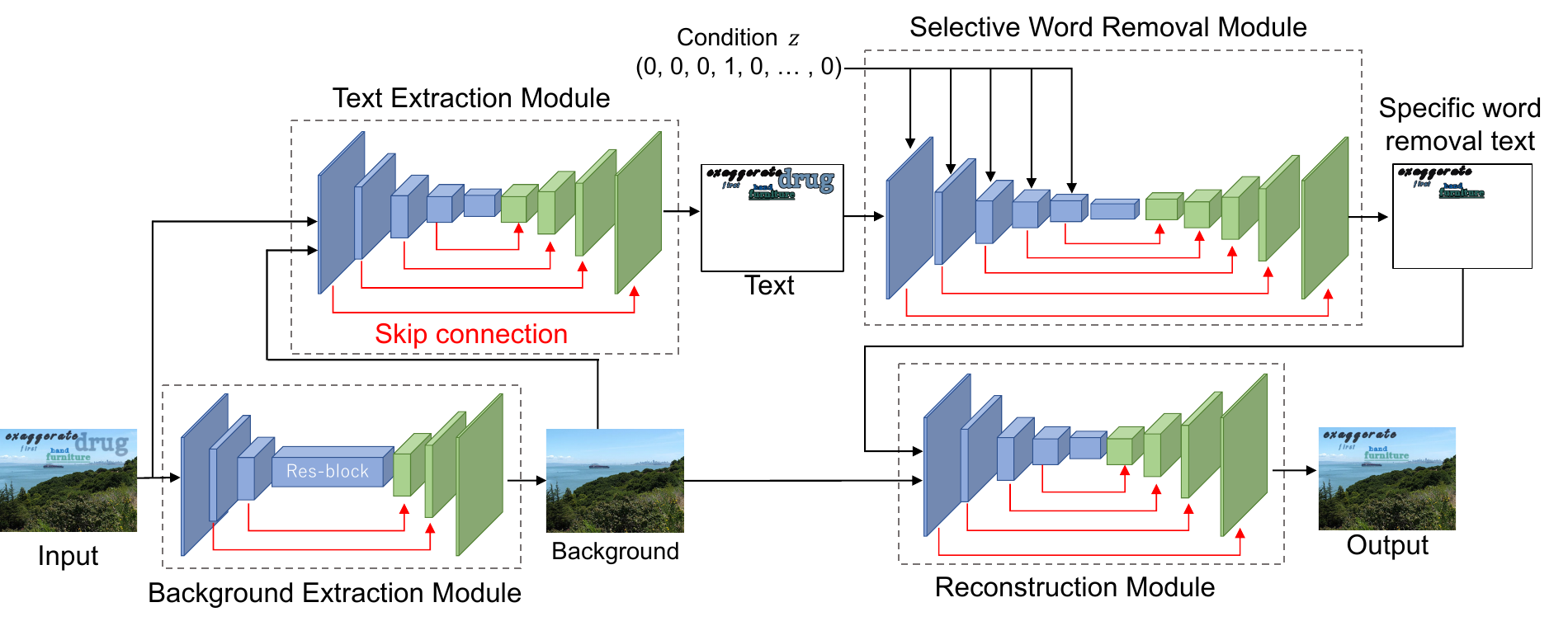}\\[-3mm]
\caption{Overview of the proposed selective scene text removal (SSTR) model. This multi-module structure makes the model training efficient. Our end-to-end SSTR model has no explicit ``word recognition'' module.}  \label{fig:overview} 
\vspace{-3mm}
\end{center} 
\end{figure}

Scene text removal (STR), or scene text eraser, is a task of removing text regions in scene images~\cite{nakamura2017scene,zhang2019ensnet,tursun2019mtrnet,zdenek2020erasing,tursun2020mtrnet++,liu2020erasenet,wang2021,tang2021,bian2022scene,lee2022}. 
As shown in Fig.~\ref{fig:task}, the conventional STR methods try to remove {\em all} text regions in scene images. This complete text removal might be reasonable for some applications. On the other hand, when we do not want to remove some important text information, a complete removal is not appropriate. For example, even if we want to remove only car license plate numbers, the existing STR methods would remove shop names from signboards as well.\par

This paper proposes a novel problem setting named {\em selective} scene text removal (SSTR) that removes target words specified by the user and does not remove the other words. Fig.~\ref{fig:task} illustrates the difference between the conventional STR and our SSTR. Note that the users need {\em not} specify the regions of the target words; they just specify the target words (like {\tt Hell} for Fig.~\ref{fig:task}), and then the proposed method automatically finds the target words, and remove them in an end-to-end manner. 
In other words, we aim to realize an automatic SSTR instead of interactive scene text removal (STR) using a bounding box tool.
SSTR will have promising applications, such as the specific removal of privacy-sensitive words or harmful words. SSTR will also make image editing~\cite{wu2019editing,Shimoda2021derendering} efficient. 
\par

The realization of SSTR has the following two difficulties. First, SSTR is a far more delicate and complicated task than STR and thus needs more training costs.  SSTR needs careful discriminability to remove various target words and to keep various non-target words. In addition, this discriminability should be held under various font styles, colors, and background images. Second, a flexible mechanism to specify the target words must be introduced. For example, even if we assume three harmful words \{{\tt drug}, {\tt cocaine}, {\tt coke}\} as candidates of the target words, we do not always want to remove all of them --- sometimes, we do not want to remove {\tt coke}. 
\par

To tackle the above difficulties, our SSTR model is designed as a multi-module structure, as shown in Fig.~\ref{fig:overview}. As detailed later, the proposed method comprises four U-Net modules with different functions: a background extraction module, a text extraction module, a selective word removal module, and a reconstruction module. This multi-module structure drastically reduces the training cost. Specifically, the proposed structure allows us  not to consider background variations in the target word removal stage. In other words, we need not think of huge combinations of words and backgrounds. Moreover, the model can be pretrained efficiently at each of the four modules independently before an end-to-end fine-tuning step. Moreover, the selective word removal module, which is the most important module for SSTR, is modeled as a conditioned U-Net so that the user can specify the target words as the test-time condition. It should be emphasized that the proposed SSTR model does not require any explicit word recognition module. We experimentally demonstrate that our SSTR removes the target words efficiently, even without a costly word recognition module. 
\par
We conduct experiments for quantitative and qualitative evaluations of the proposed model. The results on synthetic images show that the proposed model could remove target words and thus reduce their recall for detecting the target words in images from 88\% to 25\%. A comparative model based on a single U-Net could not remove the target words in most cases due to the difficulty of learning intractable combinations of target texts and their backgrounds. Another experiment with real scene images also shows the promising performance of the proposed model.\par
The main contributions of this paper are summarized as follows. First, we define a new STR task, called selective STR (SSTR), which can remove specific words and leave the other words. To the authors' best knowledge, this paper is the first attempt at SSTR. Second, we propose an efficient end-to-end neural network model with a multi-module structure that allows us to specify the target words as a test-time condition. Third, experimental results show that the proposed model can remove the target words while leaving the other words as they are. Comparison with Conditional U-Net, which is the only possible existing model that could be used as SSTR, proves the clear superiority of the proposed model.
\par

\section{Related Work\label{sec:review}}
After the first attempt of \cite{nakamura2017scene}, several STR methods have been proposed so far~\cite{zhang2019ensnet,tursun2019mtrnet,zdenek2020erasing,tursun2020mtrnet++,liu2020erasenet,wang2021,tang2021,bian2022scene,lee2022}. All these STR methods use neural network-based end-to-end trainable models. In other words, they do not use naive two-step models where text regions are first detected as bounding boxes, and then the inside of each bounding box is filled by some general inpainting methods. This may be because of three reasons. First, 
the performance of the two-step models largely depends on scene text detectors. Although recent scene text detectors (such as ~\cite{ye2020textfusenet,long2021scene,zhang2022text,liao2022real,long2022towards}) show great performance, we cannot say accurate detection is surely necessary for STR. In fact, false positives 
(i.e., regions wrongly treated as texts) are not a severe issue for STR if they can be inpainted back to the original. Second, 
bounding boxes contain many non-text pixels that need not be inpainted. End-to-end models without bounding box detection will not disturb non-text pixels. Third, end-to-end models can be incorporated into other trainable models. For example, they can be incorporated into object detectors to proactively remove text regions.\par

\begin{figure}[t]
\begin{center}
\includegraphics[width=0.85\linewidth]{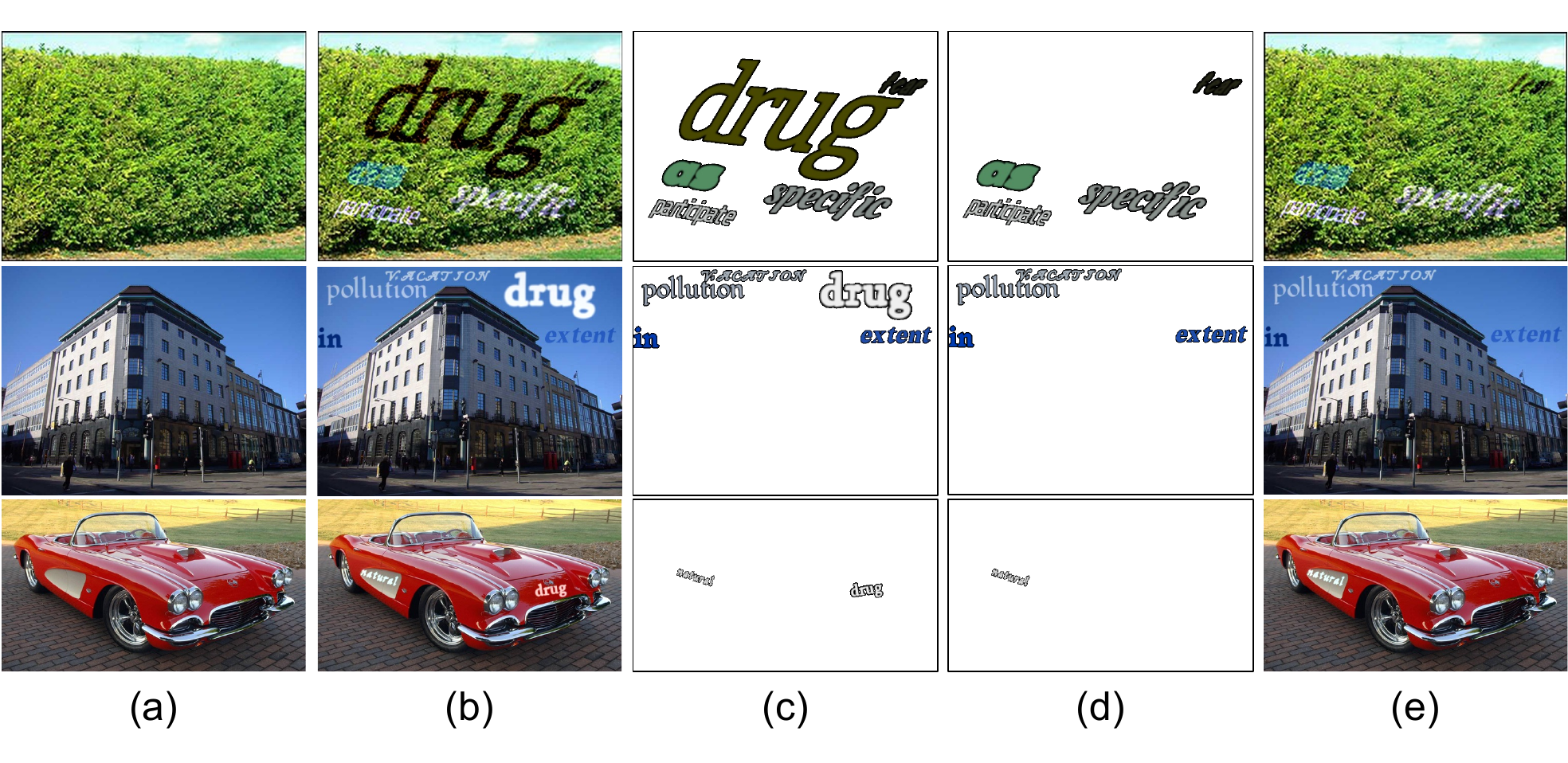}\\[-3mm]
\caption{Examples of synthetic images for training our SSTR model. (a)~Background image. (b)~Text-overlaid image by Synthtext. (c)~Overlaid text. (d)~Text image after removing the target word, {\tt drug}. (e)~Ideal output image for SSTR.}  \label{fig:synthtext}
\vspace{-3mm}
\end{center}
\end{figure}

For the success of the neural network-based STR models, scene text generators~\cite{Jaderberg14c,gupta2016synthetic,yim2021synthtiger} play a crucial role in preparing a sufficient number of training examples. Especially, SynthText~\cite{gupta2016synthetic}, which can overlay arbitrary words naturally to arbitrary natural images, has been used for training most STR models. Figs.~\ref{fig:synthtext}~(a) and (b) show three examples of SynthText. STR models can be trained using the overlaid images (b) as their input and the original (text-free) natural images (a) as the ground-truth images.\par
%
To the authors' best knowledge, SSTR is a novel task, and there is no existing SSTR model. On the other hand, it is not straightforward to modify the existing STR models for SSTR. This is simply because there is no word selection mechanism in the existing models --- their purpose is to remove all characters and words completely, and thus such a selection mechanism is unnecessary. Only MTRNet~\cite{tursun2019mtrnet} has the option to remove selected words; however, the user needs to specify the bounding boxes (i.e., the exact regions) of the target words manually. Therefore, Our SSTR model differs significantly from MTRNet because ours can automatically detect target word regions (more specifically, pixels).\par

\section{Selective Scene Text Removal (SSTR)\label{sec:methodology}}

\subsection{Overview}


In our SSTR task, we assume a set $\bm{\Omega}$ of $K$ candidate target words that can be removed from images.
For each input image $\mathbf{I}$, we specify a word
$\omega \in\bm{\Omega}$ as the target word to be removed from $\mathbf{I}$. The words in $\bar{\bm{\Omega}}\cup(\bm{\Omega}\setminus\omega)$ are not removed from $\mathbf{I}$. As an example, assume $\bm{\Omega}=\{{\tt drug}, {\tt cocaine}, {\tt coke}\}$ and $\omega= {\tt drug}$. Then, if the input image $\mathbf{I}$ contains words $\{{\tt drug}, {\tt store}, {\tt coke}\}$, {\tt drug} is removed from $\mathbf{I}$ and {\tt store} and {\tt coke} are not.
\par
As shown in Fig.~\ref{fig:overview}, the proposed model for the SSTR task comprises four modules: background extraction module, text extraction module, selective word removal module, and reconstruction module. 
These four modules are U-Net based. We used U-Net because it is still the most common image transformation model.
In the training phase, the four modules are first pretrained independently using different synthetic images created by SynthText~\cite{gupta2016synthetic}, and then fine-tuned together in an end-to-end training manner. The following sections detail individual modules.

\subsection{Four Modules for SSTR}


\textbf{The background extraction module} aims to generate a text-free image by removing all text regions from a given input image $\mathbf{I}$. Therefore, this module behaves like conventional STR, such as \cite{nakamura2017scene}. 
The basic structure of this module is a U-Net with three convolutional layers, three deconvolutional layers, and skip connections. Inspired by SRNet~\cite{wu2019editing}, four residual layers are used at the bottleneck. Both of input and output images of this module are three-channel RGB images.
The background extraction module takes a Figs.~\ref{fig:synthtext}~(b) synthetic image as its input and tries to output the corresponding Figs.~\ref{fig:synthtext}~(a) background image. Therefore, this module is trained with the MSE loss function between its output and background images. In the later experiment, we use background images prepared for the official SynthText implementation; these images are guaranteed not to contain scene text regions. 
\par
\textbf{The text extraction module} aims to extract text components from the original input images. The module is also a U-Net\cite{ronneberger2015u} with four convolutional layers, four deconvolutional layers, and 
skip connections. The input image is a six-channel image created by 
layering two three-channel RGB images of the input and background images. The output is an RGBA image (i.e., RGB plus $\alpha$-channel) showing the extracted texts. The non-text regions in the output image are treated as transparent ($\alpha=0$), and the text regions are non-transparent ($\alpha=1$) and have RGB color.  The synthetic image and its background image of Figs.~\ref{fig:synthtext}~(b) and (a) are layered as a six-channel input image, and the overlaid text image of (c) is treated as the ground-truth of the output. The MSE loss function is used to evaluate the difference between the ground-truth (c) and the actual output from the text extraction module.
\par
\textbf{The selective word removal module} is a specific module for the SSTR task and aims to remove a target word specified by the user. For example, if the target word is {\tt drug}, the module is expected to output the images like Fig.~\ref{fig:synthtext}~(d) for given images of (c). 
This module uses a Conditioned-U-Net (C-U-Net)~\cite{meseguer2019conditioned} to specify the target word $\omega$ as a condition. Specifically, the condition is represented as a $K$-dimensional one-hot vector to specify the target word $\omega$. Namely, among $K$ elements, the element corresponding to $\omega$ is set at $1$ and the others $0$.
Our C-U-Net has a FiLM (Feature-wise Linear Modulation) layer to feed the condition vector to the bottleneck part of the standard U-Net. More specifically, the output of the FiLM layer modifies the feature map at the U-Net bottleneck. The actual inputs are a text-only RGBA image like Fig.~\ref{fig:synthtext}~(c) and the condition vector. The output is also a text-only RGBA image that only shows the words in $\bar{\bm{\Omega}}\cup(\bm{\Omega}\setminus\omega)$, like (d).
\par
Finally, \textbf{the reconstruction module} outputs the SSTR result. This module is also a U-Net with skip connections.
The RGB image from the background extraction module and the RGBA image from the selective scene text removal module are layered as a seven-channel image and fed to this module as the input. Similar to the other modules, training of this module also utilizes synthetic images and the MSE loss; images like Figs.~\ref{fig:synthtext} (a) and (d) are the input images, and images like (e) are ground-truth. 

\subsection{Details of Training\label{sec:details}} 
There are no large public datasets suitable for the SSTR.
As noted above, we fully utilize synthetic scene images to train our SSTR model
We use SynthText\cite{gupta2016synthetic} to generate synthetic scene text images.
SynthText overlays words in a prespecified dictionary with various styles to scene images. This means that we know all words and their regions in the synthetic images perfectly unless the original scene images contain any scene texts. Fig.~\ref{fig:synthtext}(a)-(c) show the original image and the synthetic images by SynthText. Moreover, by removing the target word candidates as (d), we have the ground-truth images for SSTR as (e). The model is trained in an end-to-end manner after pretraining. Specifically, individual modules are first pretrained independently, and then all modules are fine-tuned in an end-to-end manner. 

\section{Experimental Results}
\subsection{Experimental Setup} 
We train the model by using the synthetic images by SynthText~\cite{gupta2016synthetic}. The word vocabulary $\bm{\Omega}$ consists of the five single-word country names, {\tt France}, {\tt China}, {\tt Germany}, {\tt Japan,} and {\tt India} ), and 500 randomly chosen words ($|\bar{\bm{\Omega}}|=500$). 
The number of fonts to print the words is 100; 25 fonts were randomly selected from each of the four font style categories (Serif, Sans-Serif, Display, and Handwriting) of Google Fonts. The original scene images without scene text, i.e., the background images, are provided with SynthText code. SynthText automatically chooses words (including a couple of country names randomly chosen from $\bm{\Omega}$) and their number, color, size, and position. We resize SynthText images to be $512\times 512$ pixels with padding to keep the original aspect ratio, and pixel values are normalized from $[0, 255]$ to $[0, 1]$ at each RGB plane.\par
SynthText provides 7,196 background images, and we use 80\% for pretraining and validation 
of the background extraction module, the text extraction module, and the reconstruction module, and 20\% for final evaluation.  
Each SynthText image contains three country names at maximum and an arbitrary number of non-country words.
Exceptionally, for pretraining the selective word removal module, we prepared another 48,000 {\em text-only} synthetic images like Figs.~\ref{fig:synthtext}~(c) and (d).  For fine-tuning all the modules in an end-to-end manner, we use the same 80\% images used for the pretraining. 
\par




Since this is the first attempt at SSTR, there are neither baseline nor state-of-the-art SSTR methods to compare with. Moreover, as reviewed in Section~\ref{sec:review}, it is not straightforward to extend most existing STR methods for SSTR.  As a barely comparable study, we attempted to use C-U-Net for SSTR. This must be the most reasonable extension of a U-Net-based STR~\cite{nakamura2017scene} to SSTR. The C-U-Net is trained to convert synthetic images like Fig.~\ref{fig:synthtext}~(b) into the resulting image (e) by itself, i.e., just by a single module. The C-U-Net model has the same structure as the selective word removal module and was trained with the same training and validation sets for fine-tuning. 

\subsection{Qualitative Evaluation} 

\begin{figure}[t]
 \begin{center}
\includegraphics[keepaspectratio, width=\linewidth]{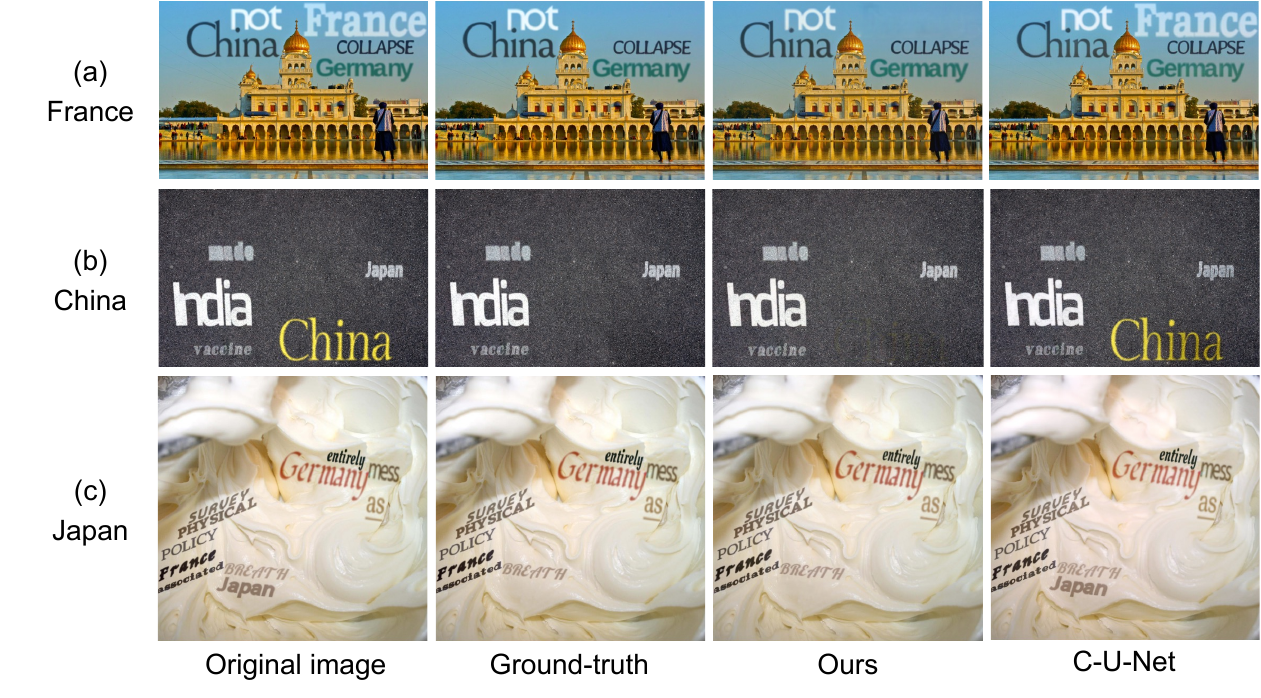}\\[-1mm]
\caption{Results by our SSTR model and the comparative model (C-U-Net).}  \label{fig:removal-result}
\vspace{-3mm}
\end{center} 
\end{figure}

\begin{figure}[t]
\begin{center}
\includegraphics[keepaspectratio, width=0.85\linewidth]{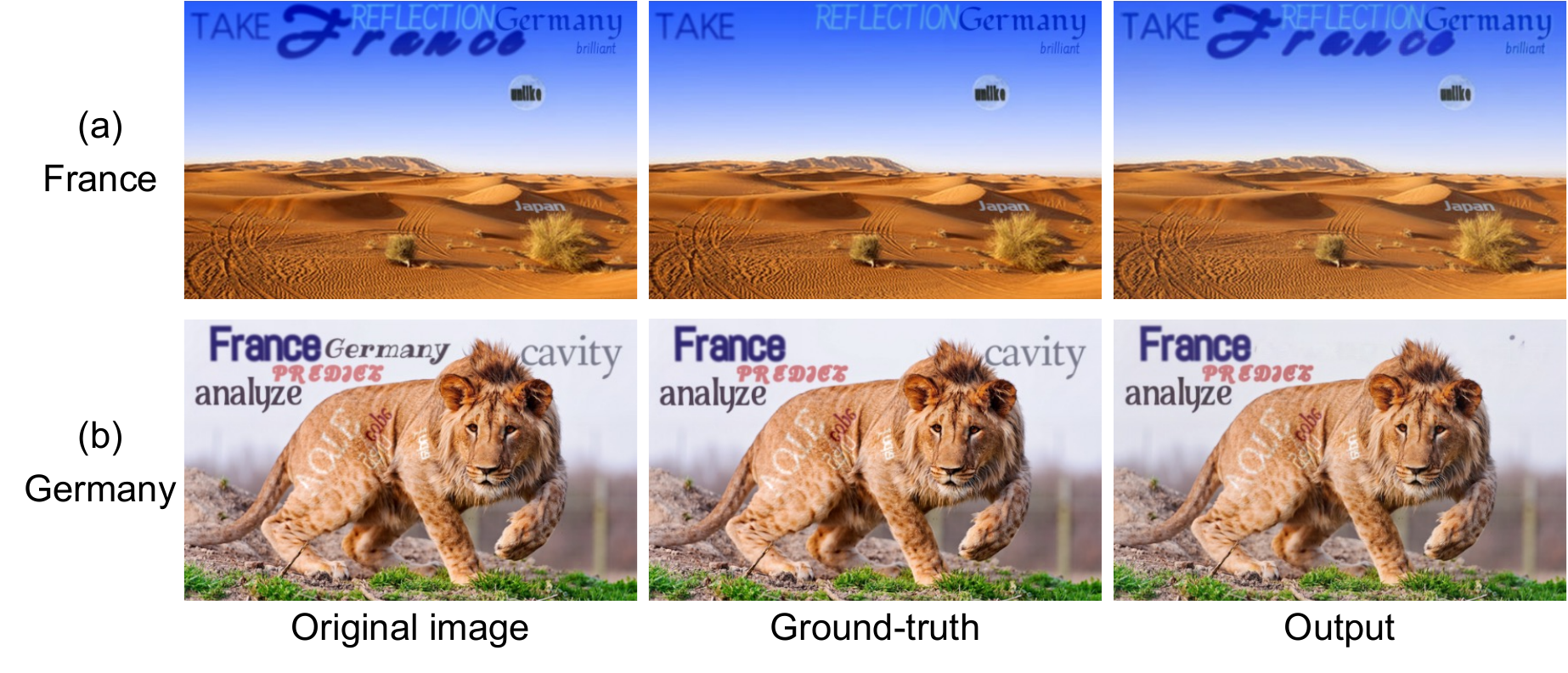}\\[-1mm]
\caption{Typical failure cases by the proposed model.}\label{fig:failure-result}
\vspace{-3mm}
\end{center} 
\end{figure}

\begin{figure}[t]
\centering
\includegraphics[keepaspectratio, width=0.97\linewidth]{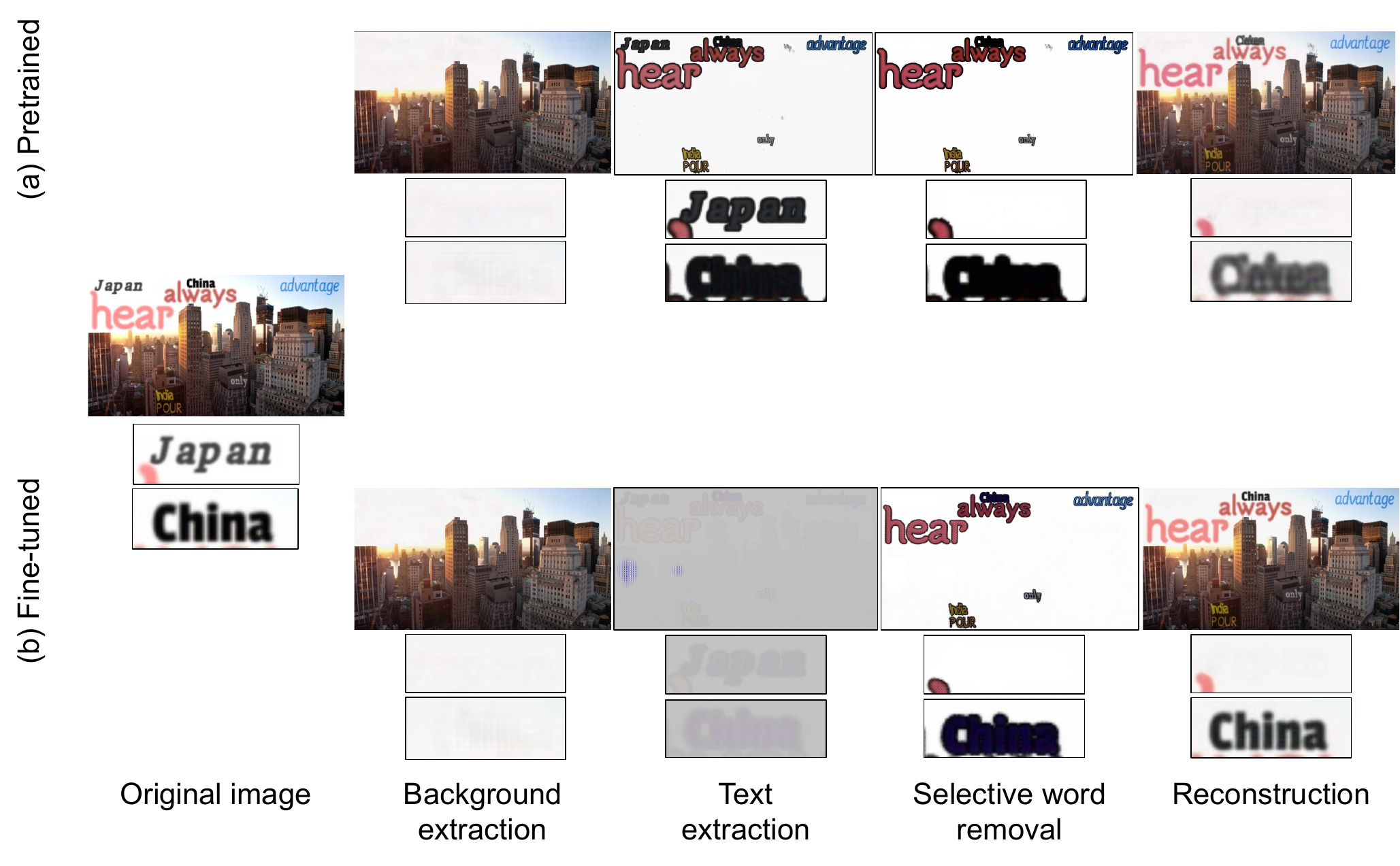}
\caption{Outputs from each module (a)~before fine-tuning (i.e., right after pretraining) and (b)~after fine-tuning. The target word is {\tt Japan} in (a), (b). For (a) and (b), closer views of the target word {\tt Japan} and a non-target word {\tt China} are attached below each output result.}\label{fig:module-result}
\end{figure}

Fig.~\ref{fig:removal-result} shows SSTR results on test images by our model and the comparative model. First, our model shows successful SSTR results by removing the target word accurately regardless of its size, color, location, and style, and then inpainting the removed region. Second, our model does not remove the non-target word. In (a), for example, the non-target word {\tt China} is left with its original appearance, while the target word {\tt France} is completely removed.
\par
On the other hand, the comparative model could not remove the target word in most cases. Its output image is near-identical to the original image. This indicates that the comparative model fell into a sub-optimal solution to output the input image almost as is. In a sense, it is a reasonable solution for the comparative model; rather than trying to remove small target word regions with the risk of unexpected side effects, the comparison model returns images that are nearly identical to the input to reduce MSE loss.\par

Fig.~\ref{fig:failure-result} shows typical failure cases. In (a), the target word {\tt France} is printed in an uncommon style, which might be far different from the trained styles. Our model could not find it as the target word and output an image that is nearly identical to the original image. In (b), the target word {\tt Germany} is removed successfully; however, a non-target word {\tt cavity} is also removed erroneously. Such a false positive happens when a non-target word is somewhat similar to the target word. \par
Fig.~\ref{fig:module-result} shows the internal outputs from individual modules before and after fine-tuning. All the modules work almost properly, and the target word {\tt Japan} is removed in (a) even before the fine-tuning. However, a closer look at the reconstructed image in (a) reveals that colors and words are slightly faded and blurred, respectively. In particular, focusing on the non-target words, {\tt China} in (a) is very blurred. 
On the other hand, the output of the text extraction module in (b) has changed color as a result of fine-tuning, but this does not affect the processing of the subsequent modules. 
In fact, the reconstructed image in (b) shows that the target word {\tt Japan} has been removed, while the rest of the image is almost the same as the original image.
This suggests that by fine-tuning, the proposed model has learned to modify its internal output so that it is better suited for the final output.

\subsection{Quantitative Evaluation\label{sec:quantitative}} 
Following the past STR papers (e.g., \cite{zhang2019ensnet}), we use recall of the target word after removal and general image quality metrics (by PSNR and MSE) for quantitative evaluation. The (area-based) recall value is calculated by using the overlap ratio between the ground-truth region and the bounding box by a state-of-the-art scene text detector, TextFuseNet~\cite{ye2020textfusenet}. On the test images, TextFuseNet achieves 88.15\% for recall and 73.51\% for precision.


Table~\ref{table:performance} shows the quantitative evaluation results. The \textcolor{magenta}{highlight} is that the proposed model could decrease the recall from 88.15\% to 24.82\% at the target word regions. This lower recall proves that the proposed model could remove the target words successfully in most cases. In contrast, the comparative method, C-U-Net, could not decrease the recall at all. As we emphasized in this paper, SSTR is much more difficult than STR because SSTR needs to solve the difficult problem of discriminating the target words from non-target words on various background images. Therefore, C-U-Net cannot deal with the difficulty by itself. Our SSTR model is free from background variations in its selective word removal stage and could achieve better removal performance.
PSNR and MSE also confirm this performance difference at the target word regions; both metrics prove that the proposed model is far closer to the ground-truth than C-U-Net.
\par
In the non-target word regions, the proposed model achieved almost the same recall as the original. This means that the proposed model did not exhibit the side-effect of removing non-target words when removing target words. In other words, the proposed model realizes an accurate selection of target words from all words in the input image without any explicit word classification module.\par
The image quality evaluations by PSNR and MSE in the non-target word and background regions show that C-U-Net achieved slightly better quality than the proposed model. This is because C-U-Net fell into a sub-optimal solution to output the input image almost as is. In contrast, the proposed model undergoes slight color changes during its SSTR process. The difference between the two models, however, is not very significant, and therefore, the image quality of the proposed model is also still high enough.

\begin{table}[t]
\center
\caption{Quantitative evaluation result. R:~Recall(\%). SN:~PSNR. M:~MSE(\%).
\label{table:performance}}
{
\begin{tabular}{l|rrr|rrr|rr}\hline
  & \multicolumn{3}{c|}{Target word} & \multicolumn{3}{|c}{Non-target word} & \multicolumn{2}{|c}{Background}\\ \cline{2-9}
 & 
\multicolumn{1}{|c}{R$\downarrow$} & 
\multicolumn{1}{c}{SN$\uparrow$} & 
\multicolumn{1}{c}{M$\downarrow$} & 
\multicolumn{1}{|c}{R$\uparrow$} & 
\multicolumn{1}{c}{SN$\uparrow$} & 
\multicolumn{1}{c}{M$\downarrow$} &
\multicolumn{1}{|c}{SN$\uparrow$} & 
\multicolumn{1}{c}{M$\downarrow$}  
 \\ \hline
      Proposed & \textbf{\textcolor{magenta}{24.82}} & \textbf{23.32} & \textbf{1.78} & \textcolor{magenta}{86.88} & 30.13 & 0.19 & 32.09 & 0.08 \\
          C-U-Net & \textcolor{magenta}{88.15} & 14.57 & 5.23 & \textbf{\textcolor{magenta}{87.76}} & \textbf{31.61} & \textbf{0.17} & \textbf{38.62} & \textbf{0.02}\\
        \hline  
\end{tabular}}
\end{table}

\subsection{Comparison with Two-step Model}
\begin{figure}[t!]
\begin{center}
\includegraphics[keepaspectratio, width=0.88\linewidth]{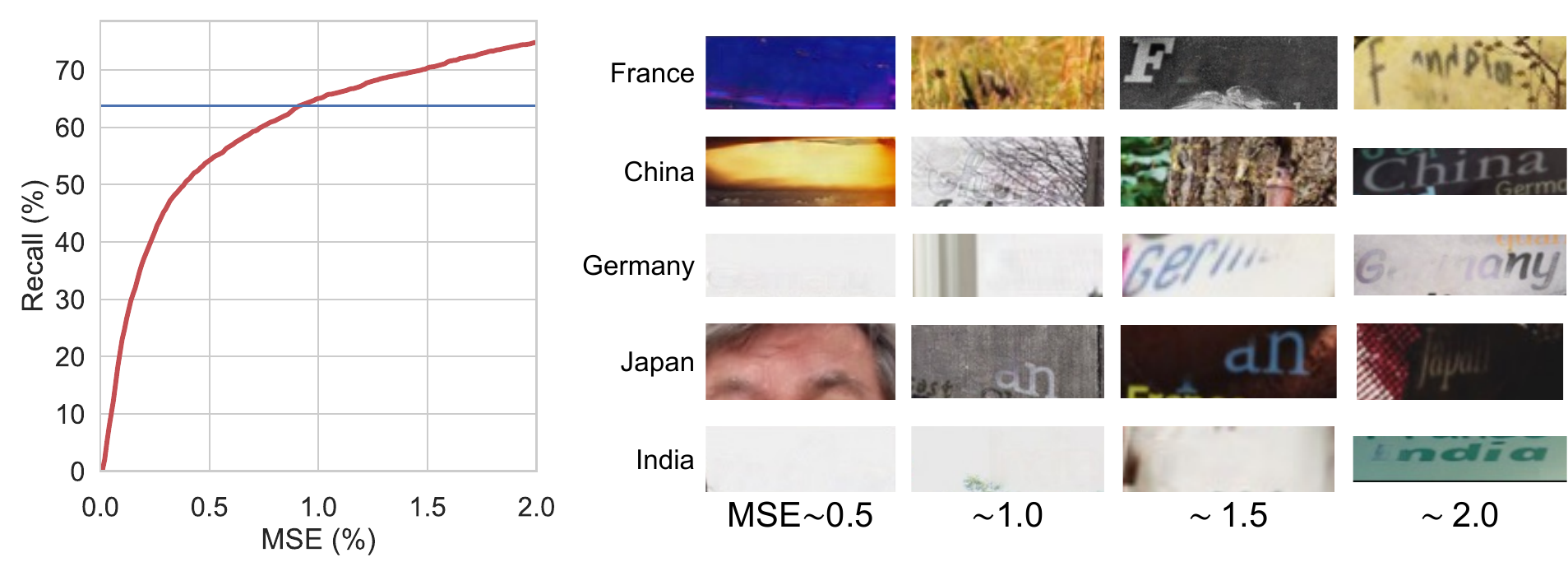}\\[-3mm]
\caption{Comparison with two-step model. Left: Recall (\%) of the proposed model as a function of the threshold for MSE. The horizontal line indicates the recall (63.76\% ) by  SwinTextSpotter~\cite{huang2022swintextspotter}. Right: Removal examples at different MSE values.}\label{fig:two-step}
\vspace{-3mm}
\end{center} 
\end{figure}
One might think that using a scene-text detector followed by an inpainter is more efficient for realizing SSTR. However, this two-step model is not efficient. 
More specifically, our model internally achieves the same target word detection performance with fewer computations than the two-step model. 
\par
Here we compare our model to a two-step model with SwinTextSpotter~\cite{huang2022swintextspotter}, which is a state-of-the-art method to detect and recognize scene texts.
SwinTextSpottor achieves a recall rate of 63.76\% for the target word on our SynthText image set. Since our model does not detect the target word explicitly, we measure MSE between the background image and the resulting image of the target word region. If MSE is zero, it is natural 
to consider that our model detects the target word successfully and removes it completely. If MSE is non-zero but still smaller than a small threshold, our model detects the target word and removes it almost completely. Fig.~\ref{fig:two-step} shows a detection rate (i.e., recall)  of our model on the SynthText dataset as a function of the threshold. When the threshold MSE equals $1\%$, our model achieves a recall of 65.04\%. The example images in Fig.~\ref{fig:two-step} prove that the target words are almost invisible when MSE $\sim 1\%$. Consequently, the recall rate of our model is approximately 65.04\%, which is almost the same as SwinTextSpottor.  (For a fair comparison with SwinTextSpottor, this recall rate is measured in this MSE-based way and thus different from the recall rate in Table~\ref{table:performance}, where the rate is evaluated in the standard way.) For computational costs, SwinTextSpotter
 requires $0.697$ seconds per SynthText image on an NVIDIA Titan RTX GPU, whereas our model requires $0.104$ in total. If we add the computation time of the inpainter, this difference becomes larger. \par    
\subsection{SSTR on Real Scene Texts\label{sec:real}} 

\begin{figure}[t!]
\begin{center}
\includegraphics[keepaspectratio, width=0.95\linewidth]{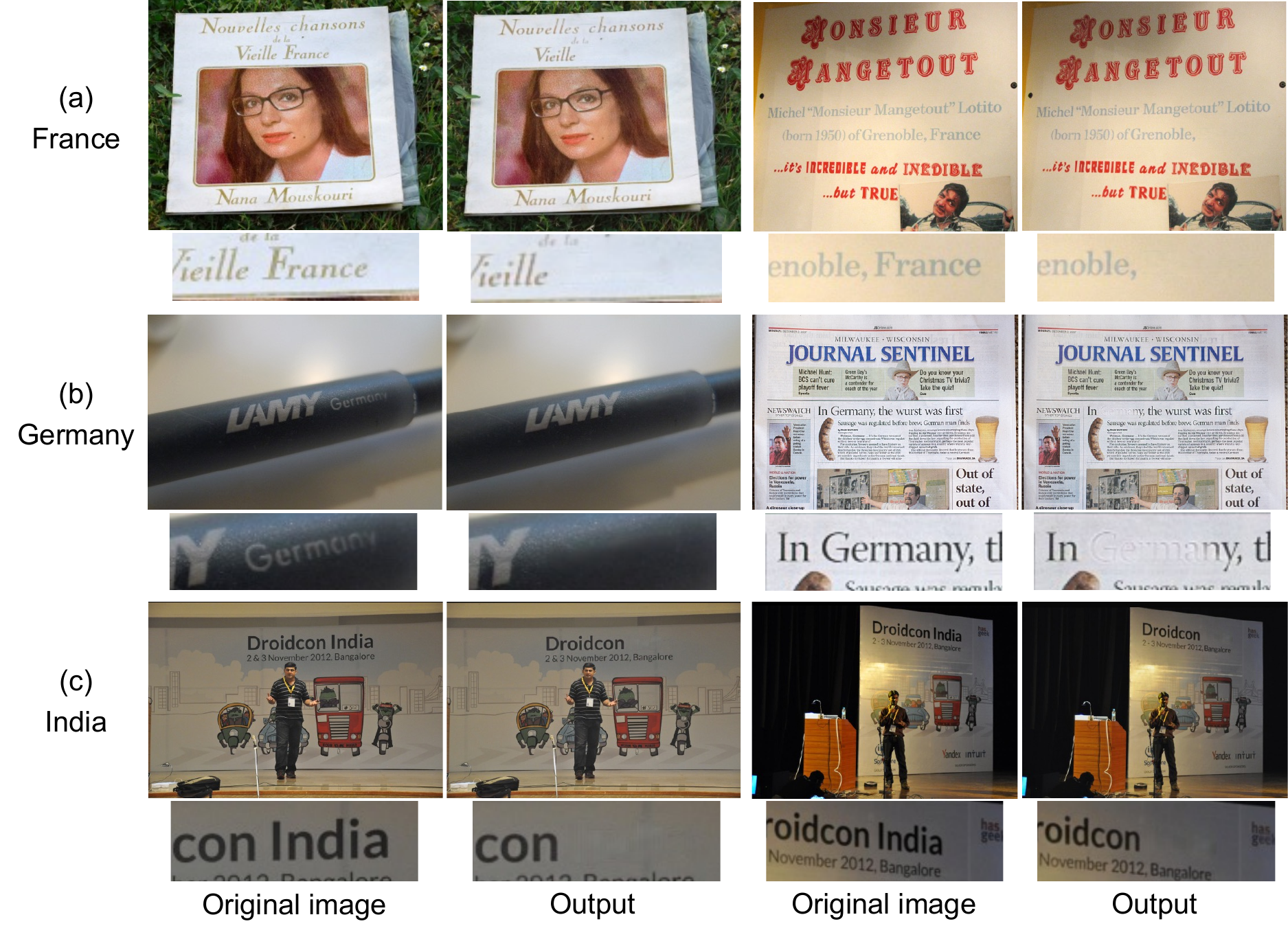}\\[-3mm]
\caption{SSTR results on real images from OpenImages v4.}\label{fig:real-result}
\vspace{-3mm}
\end{center} 
\end{figure}

Finally, we apply the proposed model to real-scene text images. Since no public scene text dataset is appropriate for the SSTR task, we evaluate the performance of the proposed model qualitatively using 50 scene text images (10 for each country name) from OpenImages v4 dataset.
Fig.~\ref{fig:real-result} shows the application results. Although we trained the model with synthetic images, it still can remove the target words accurately. In the right example of (b), however, the model cannot remove the target word {\tt Germany} completely. Our visual inspections turn out that small word regions are difficult to be removed. This might be because that small words in real images are often more blurry and/or distorted by JPEG compression and thus have a domain gap from those in synthetic images. On the other hand, larger word regions are removed more accurately. According to our visual inspection, the successful removal rate was $76\%$ ($=38/50$). \par

\section{Conclusion and Future Work}
This paper proposed a new task called selective scene text removal (SSTR), which aims to remove a specific word from an input image instead of all words. We also proposed a neural network-based model for SSTR. The model is organized in a multi-module structure to pretrain individual modules in an efficient way. We confirmed that the proposed model trained by synthetic images performs well not only with synthetic images but also with real images, whereas a comparative model in a single-module structure cannot work even with synthetic images.\par
As noted in Section~\ref{sec:real}, one limitation of the current trial is that it shows the difficulty in removing small texts. A possible remedy is to employ a more accurate scene text synthesis method to simulate actual distortions on small texts. Another limitation of the current trial is that we still do not fully utilize the end-to-end nature of the proposed model. We can utilize it by embedding the proposed model into another neural network-based model for a specific purpose, such as scene text editing and object detection. Finally, we would like to extend our model to remove synonyms and/or relevant words; for example, specifying the target word {\tt dog} would remove words related to dogs, such as {\tt beagle} and {\tt bulldog}. Conditioning by a semantic vector of {\tt dog} (instead of the one-hot vector) will help this extension.
%
\section*{Acknowledgment}
This work was supported in part by JSPS KAKENHI Grant Numbers JP22H00540.

\bibliography{BMVC_Mitani_camera-ready}
\end{document}